%% file: main.tex
\documentclass[runningheads]{llncs}
\usepackage[utf8]{inputenc}

\usepackage{cite}
\usepackage{amsmath,amssymb,amsfonts}
\usepackage{algorithmic}
\usepackage{graphicx}
\usepackage{textcomp}
\usepackage{xcolor}
\usepackage{ulem}
\usepackage{algorithm2e}
\usepackage{authblk}

\def\BibTeX{{\rm B\kern-.05em{\sc i\kern-.025em b}\kern-.08em
    T\kern-.1667em\lower.7ex\hbox{E}\kern-.125emX}}
\begin{document}

\title{Semantic Segmentation for Real-World and Synthetic Vehicle's Forward-Facing Camera Images\\
}

\titlerunning{Semantic Segmentation For Vehicle's Forward-Facing Camera Images}

\authorrunning{Tuan T. Nguyen et al.}
\date{August 2021}

\author{Tuan T. Nguyen$^\dagger$ \inst{1}\and
Phan Le \inst{2} \and
Yasir Hassan\inst{3}\\
\and Mina Sartipi\inst{4}}

\institute{Center of Urban Informatics and Progress\\
  University of Tennessee at Chattanooga \\
  615 McCallie Ave, Chattanooga TN 37403\\ 
  \email{xwz778@mocs.utc.edu}\inst{1} 
  \email{bbz181@mocs.utc.edu}\inst{2} \email{krs824@mocs.utc.edu}\inst{3} \email{mina-sartipi@utc.edu}\inst{4}
 }
  



\maketitle
\input{01-abstract}
\input{02-Introduction}
\input{03-RelatedWork}

\input{04-Method}
\input{05-Experience}
\input{06-Conclusion}
\bibliographystyle{splncs04}
\bibliography{sn-bibliography}
\bibliography{main}

\end{document}

%% file: 01-abstract.tex
\begin{abstract}
In this paper, we present the submission to the 5th Annual Smoky Mountains Computational Sciences Data Challenge - Challenge 3. This is the solution for semantic segmentation problem in both real-world and synthetic images from a vehicle’s forward-facing camera. We concentrate in building a robust model which performs well across various domains of different outdoor situations such as sunny, snowy, rainy, etc. In particular, our method is developed with two main directions: model development and domain adaptation. In model development, we use the High Resolution Network (HRNet) as the baseline. Then, this baseline's result is processed by two coarse-to-fine models: Object-Contextual Representations (OCR) and Hierarchical Multi-scale Attention (HMA) to get the better robust feature. For domain adaption,  we implement the Domain-Based Batch Normalization (DNB) to reduce the distribution shift from diverse domains. Our proposed method yield 81.259 mean intersection-over-union (mIoU) in validation set. This paper studies the effectiveness of  employing real-world and synthetic data  to handle the domain adaptation in semantic segmentation problem.

\end{abstract}

%% file: 02-Introduction.tex
\section{Introduction}
The goal of semantic segmentation is assigning an object label to each pixel of an image. It is one of the essential topic in computer vision because of its critical application for important practical tasks such as autonomous driving. Due to recent development of deep learning, significant improvement have been made in semantic segmentation. One of the most successful breakthroughs is learning robust features from well-designed segmentation frameworks. The performance of segmentation model is measured by mIoU (Mean Intersection-Over-Union). As an example, FCN \cite{long2015fully} can achieves 65.3\% mIoU on the test set of Cityscapes \cite{cordts2016cityscapes}. This score is significantly improved by HRNet \cite{sun2019high} and reached 81.6\% mIoU. Additionally, coarse-to-fine models such as OCR \cite{yuan2020object} and HMA \cite{tao2020hierarchical}  have been developed to take the predicted segmentation features (coarse) as input and generate better features (fine), e.g., the combination of OCR \cite{yuan2020object} and HRNet \cite{sun2019high} yield 84.2\% mIoU on Cityscape test set, this score is boosted to 85.4\%  with the combination of HMA \cite{tao2020hierarchical}, OCR \cite{yuan2020object} and HRNet \cite{sun2019high}. 

\begin{figure}[h]
    \centering
    \includegraphics[scale=0.35]{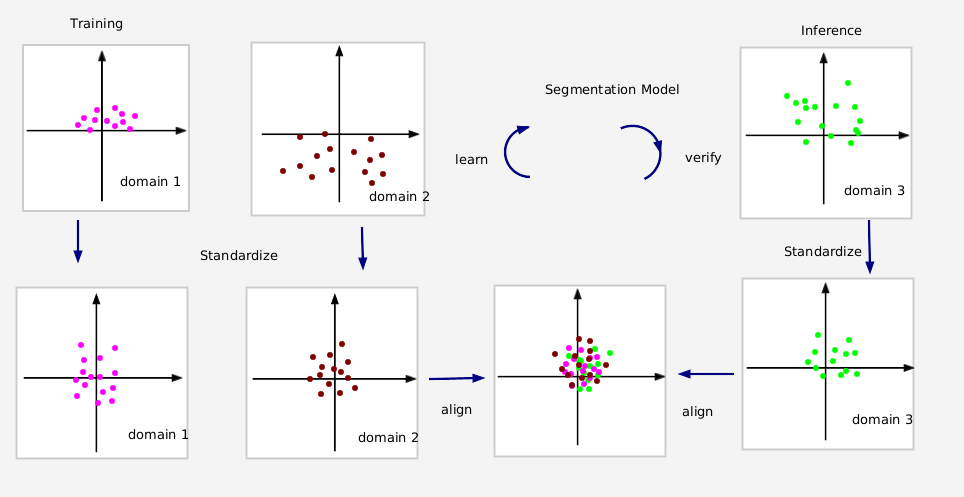}
    \centering 
    \caption{Illustration of the proposed Domain-based formulation, motivated by \cite{zhuang2020rethinking}. The data from different distribution is standardized in training process with DBN to generate a robust generalized feature. \textit{This figure is best viewed in color}}
    \label{fig:dbn}
\end{figure}

However, Data hunger is one of the main drawbacks of the current state-of-the-art models. They often require a large amount of pixel-level labeled images to generate a reliable result. In addition, these data-driven based models usually perform poorly in unseen image domains. For example, the model trained on data from a sunny day will perform poorly on the data from a snowy day. Therefore, the problem occurs because annotating pixel-level labeled image is costly in term of labor and time. For instance, it takes 7,500 hours of manual work to annotate the segmentation ground truth to produce the Cityscapes dataset \cite{cordts2016cityscapes}. One solution for this problem is utilizing the huge easy-to-get synthetic data, e.g., CARLA \cite{dosovitskiy2017carla}, SYNTHIA \cite{ros2016synthia} and GTA5 \cite{richter2016playing}. In this case, we need to handle the problem of distribution shift by aiming at transferring the knowledge from a simulation domain to a real-world domain. In this paper, we implement the Domain-Based Batch Normalization (DBN) to align data from different distribution and generate a better robust feature, which is shown  in Figure \ref{fig:dbn}. This domain adaptation technique is motivated by the Camera-Based Batch Normalization \cite{zhuang2020rethinking}. 

For the Challenge 3 of the $5^{th}$ Annual Smoky Mountains Computational Sciences Conference (SMC), we are focusing on solving the problem of distribution shift which occurs when the semantic segmentation model is trained on source domain (e.g., synthetic data, and sunny weather) but tested on unseen target domain (e.g., real-world and weather condition that is not appear on source domain). The provided training/validation dataset contains a total of 5600 images, divided into 700 images from the training set of Cityscapes \cite{cordts2016cityscapes}, and the remaining images are synthetic images from CARLA \cite{dosovitskiy2017carla}, which represent seven different weather conditions. The testing set is not provided but it comes from 3 sources: the subset of  Cityscapes' testing data, synthetic images from different weather conditions and an unknown data source. Our goal is to build a robust semantic segmentation model that can handle the distribution shift and perform accurately on both real-world and synthetic images from diverse weather situations. We achieve this goal by developing a model with two main purposes:

\begin{enumerate}
    \item Model development: building the robust semantic segmentation framework.
    \item Domain adaptation:  developing technique to reduce the distribution shift of diverse data distributions.
\end{enumerate}

For model development, we combine HRNet \cite{sun2019high}, OCR \cite{yuan2020object} and HMA \cite{tao2020hierarchical} to construct a semantic segmentation framework. For Domain Adaptation, we implement a DBN module which standardize the data from a different distribution. The details of implementation for all of them will be clarified in section \ref{sec:method}. Our method is evaluated on the validation set and the result will be shown in section \ref{sec:experience}.

%% file: 03-RelatedWork.tex
\section{Related Work}
\label{sec:related_work}
A significant amount of current state-of-the-art works in the field of semantic segmentation are based on the use of deep-learning algorithms to get object labels for the input pictures. Such approaches, depending on the model architecture, including, but are not limited to the usage of Convolutional Neural Networks (CNNs), multi-scale architecture, encoders and decoders, and, more recently, dilated convolutions. Some of the works published under the topic of semantic segmentation also concentrate on post-processing of pre-existing models results to improve their accuracy. In this section, we will briefly discussed some main works on semantic segmentation topics.

    
     \textbf{High resolution network:} The main categories of features in  computer vision deep feature learning are low-resolution features for image classification task and high-resolution features for other tasks (e.g., object detection or human pose estimation and semantic segmentation). In order to compute the high-resolution features, there are two main directions: The first one is computing the low-resolution features and transfer from this learnt low-resolution features to high-resolution features, e.g., Encoder-Decoder \cite{peng2016recurrent}, Hourglass \cite{newell2016stacked}. The second one is remaining high-resolution features and enhancing them with the low-resolution convolutions, e.g., HRNet \cite{sun2019deep, sun2019high}, GridNet \cite{fourure2017residual}.
    
     \textbf{Multi-scale context} 
    PSPNet~\cite{Zhao_2017_CVPR} employs a spatial pyramid pooling module\cite{he2015spatial} to capture multi-scale context. DeepLab~\cite{chen2018encoder} utilizes Atrous Spatial Pyramid Pooling (ASPP), resulting in a denser feature than PSPNet. Recently, ZigZagNet~\cite{lin2019zigzagnet} and ACNet~\cite{fu2019adaptive} have created multi-scale contexts by leveraging intermediate features in additional to the ones from the network trunk's last layer.
    
    \textbf{Relational context} 
    While pyramid pooling methods focus on fixed and square regions, relational context methods focus on the relationship between pixels to build contextual representations based on image composition~\cite{tao2020hierarchical}. This was utilized by  DANET~\cite{fu2019dual}, CFNet~\cite{zhang2019co}, OCRNet~\cite{yuan2020object}, and other similar works \cite{zhang2019acfnet, chen2019graph, chen2018a2nets, liang2018symbolic} to construct better context.

\begin{figure}[h]
    \centering
    \includegraphics[scale=0.17]{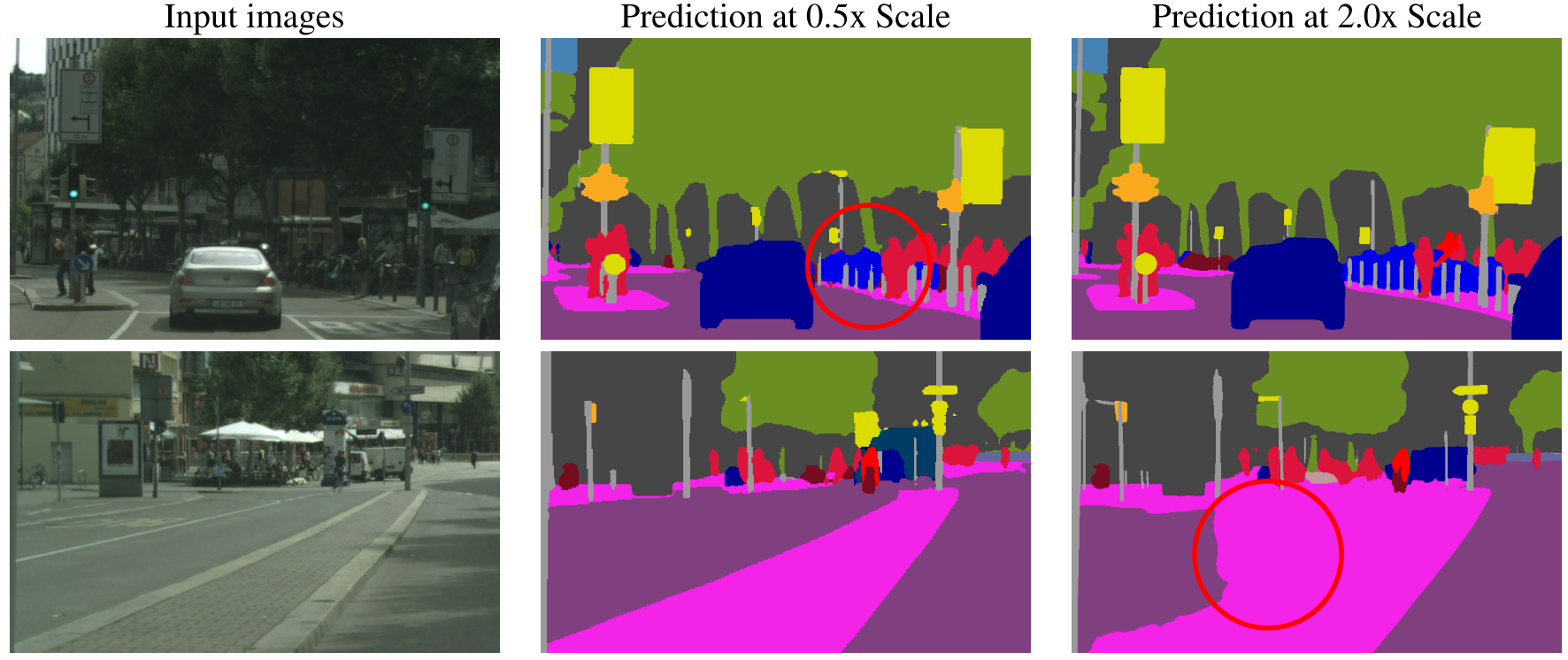}
    \centering 
    \caption{Visualization of failures for semantic segmentation relating to inference scale from \cite{tao2020hierarchical}. In the first row, the thin posts are performed well on low resolution image (0.5x) but better predicted in high resolution image (2.0x). In the second row, the large object (road) is worse predicted at higher resolution (2.0x)}
    \label{fig:hma}
\end{figure}

    \textbf{Multi-scale inference} 
They are coarse-to-fine models which take the pixel-level segmentation representation from other model (coarse) to generate a better pixel-level representation (fine). The motivation comes from the observation that the model trained on high resolution image perform better on the thin structures and the edges of objects while the model trained on low-resolution image predicted better on large structure. This observation is shown in Figure \ref{fig:hma}. Then, a pixel-wise attention mask is learnt to aggregate the pixel-level segmentation representation of low-resolution images and high-resolution images. Examples of important works in this field are \cite{chen2016attention, yang2018attention, tao2020hierarchical}.

%% file: 04-Method.tex
\section{Methodology}
\label{sec:method}

\begin{figure}[h]
    \centering
    \includegraphics[scale=0.27]{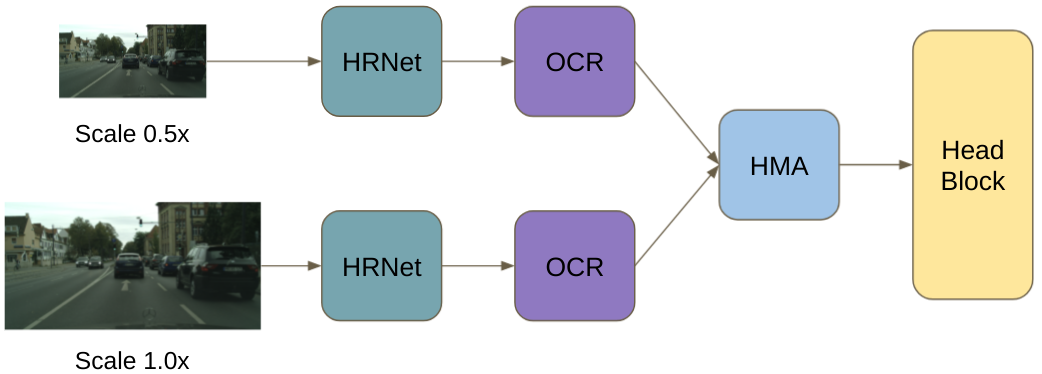}
    \centering 
    \caption{\textbf{Method Architecture}. The DBN is implemented by replacing BN layers inside all blocks with DBN layers. The detail of blocks are shown in: figure \ref{fig:hrnet} - HRNet block, figure \ref{fig:ocr2} - OCR block, figure \ref{fig:hma2} - HMA block.} 
    \label{fig:arch}
\end{figure}

In our method, the images and ground truth segmentation model are fed to HRNet \cite{sun2019high} to get the coarse pixel-level representation. Then, this coarse pixel-level representation is processed by OCR  \cite{yuan2020object} to learn a better object-contextual representation for each pixel. After that, the HMA \cite{tao2020hierarchical} take this pixel-level object-contextual representations from two scaled images (0.5x and 1.0x) as input and generate the final features by pixel-wise weighted aggregation with learnt attention masks. Within all architectures, the normal batch normalization layers is replaced by DBN. All of the architectures are shown in Figure \ref{fig:arch}.

The rest of this section are presented in three subsections. First, we are describing HRNet  \cite{sun2019high} in section \ref{sec:hrnet}. Then, two coarse-to-fine models are presented in Sections \ref{sec:ocr} (OCR\cite{yuan2020object}) and \ref{sec:hma} (HMA\cite{tao2020hierarchical}). Finally, the domain adaptation module DBN is conducted in section \ref{sec:dbn}.

\subsection{High-Resolution Representations Network (HRNet)}
\label{sec:hrnet}

\begin{figure}[h]
    \centering
    \includegraphics[scale=0.18]{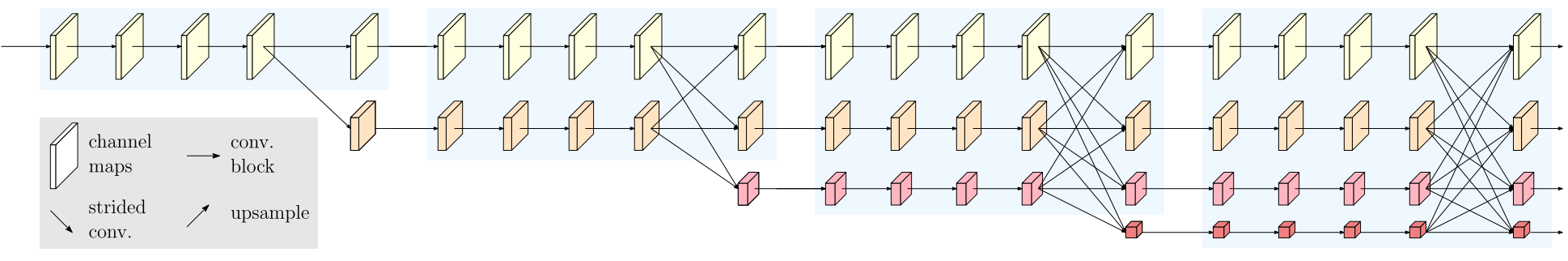}
    \centering 
    \caption{Visualization of High-resolution network architecture from \cite{sun2019high}. There are four stages. The 1st stage consists of high-resolution convolutions. The 2nd (3rd, 4th) stage repeats two-resolution (three-resolution, four-resolution) blocks.}
    \label{fig:hrnet}
\end{figure}

High-resolution representation network was first invented to handle the problem of human pose estimation \cite{sun2019deep}. Then, it was modified to better address semantic segmentation problem \cite{sun2019high}. This network remains the high-resolution representation throughout the training process and the learnt feature is strengthened by performing high-to-low resolution convolutions in parallel. The network architecture is shown in Figure \ref{fig:hrnet}. There are four stages in this architecture. The first one is a normal high-resolution convolutions. The second (third, fourth) stage repeats two-resolution (three-resolution, four-resolution) blocks. In multi-resolution blocks, firstly, the down-sampling technique  is executed on the input channel to get several channel's subset; then, the regular convolution process is performed for each subset of different spatial resolution. In general, it mirrors the full-connection behavior of the regular convolution.

\begin{figure}[h]
    \centering
    \includegraphics[scale=0.17]{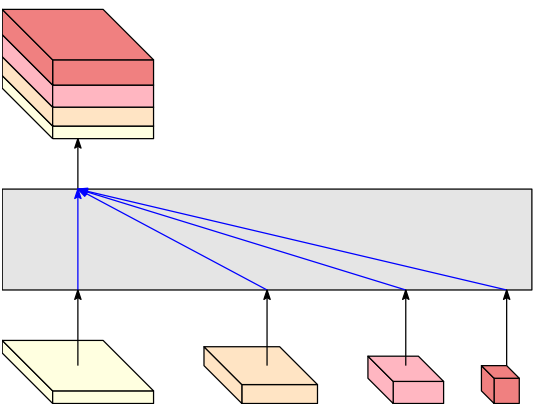}
    \centering 
    \caption{Visualization of generating final representation in HRNet from \cite{sun2019high}.  After the low-resolution feature is rescaled, all feature is concatenated to generate the final feature. }
    \label{fig:hrnet2}
\end{figure}

The final feature was created by combining the representation from all resolutions. The combining process includes rescaling the low-resolution representations with upsampling to the high resolution and concatenating all the subset of representations. This process is illustrated in \ref{fig:hrnet2}.

\subsection{Object-Contextual Representations}
\label{sec:ocr}
 
\begin{figure}[h]
    \centering
    \includegraphics[scale=0.27]{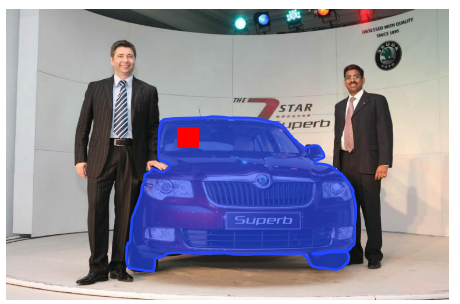}
    \centering 
    \caption{OCR visualization from \cite{yuan2020object}. The context of \textcolor{red}{red} pixel is expected to be a set of pixels within the \textcolor{blue}{blue}-marked object (Car) that this red pixel is belong to.}
    \label{fig:ocr}
\end{figure}

In OCR \cite{sun2019high} block, the context aggregation technique is applied on the output of HRNet block. This technique was derived from the idea that the label of the pixel is the category of the object that the pixel belongs to. Thus, the pixel-level representation can be improved by aggregating the representations of pixels that  belong to the same object. This process is shown in Figure \ref{fig:ocr}

\begin{figure}[h]
    \centering
    \includegraphics[scale=0.25]{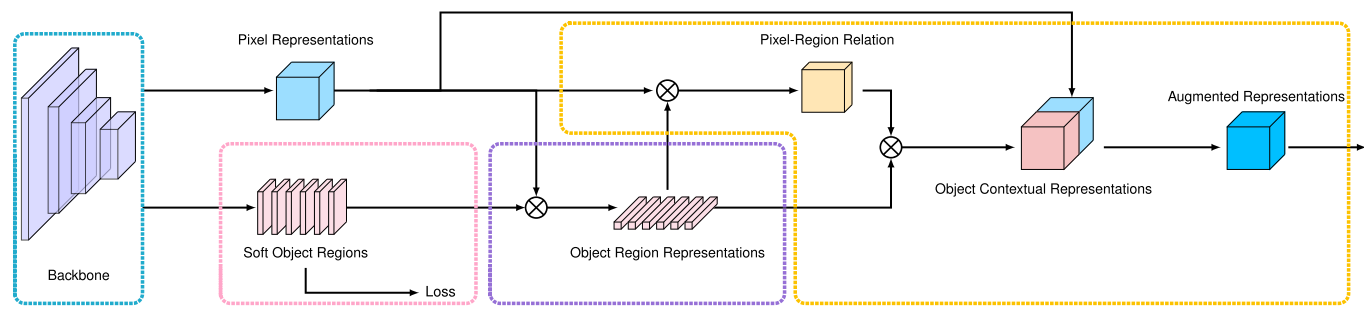}
    \centering 
    \caption{Illustrating of OCR's pipeline from \cite{yuan2020object}. (1) Assigns all the pixels in a image into K soft object region in the \textcolor{pink}{pink} dashed box. (2)  Calculate K object region representation in the \textcolor{violet}{violet} dashed box. (3) Calculate the new representation for each pixel in the \textcolor{orange}{orange} dashed box.}
    \label{fig:ocr2}
\end{figure}

The OCR's formulation includes 3 steps as described below. This pipeline is illustrated in Figure \ref{fig:ocr2}.

\begin{enumerate}
    \item Assign all  pixels in an image into K soft object region. This is conducted by building a small supervised learning model using the pixel-level representation from HRNet block as input and the ground-truth segmentation as labels.
    \item Calculate K object region representation by aggregating the representations of all the pixels in the corresponding object region.
    \item Calculate the new representation for each pixel by weighted aggregating the K object region representations. The weights are the probability distribution that this pixel belongs object regions.
\end{enumerate}

\subsection{Hierarchical Multi-scale Attention Model (HMA)}
\label{sec:hma}

Hierarchical Multi-scale Attention Model (HMA) is a multi-scale inference model which aggregates the segmentation results from low-dimension images and high-dimension image to generate a better segmentation output. The motivation comes from the observation that the model trained on high resolution image performs better on the thin structures and the edges of objects while the model trained on low resolution image predicts better on large structure. This is visualized in figure  \ref{fig:hma}. 

\begin{figure}[h]
    \centering
    \includegraphics[scale=0.17]{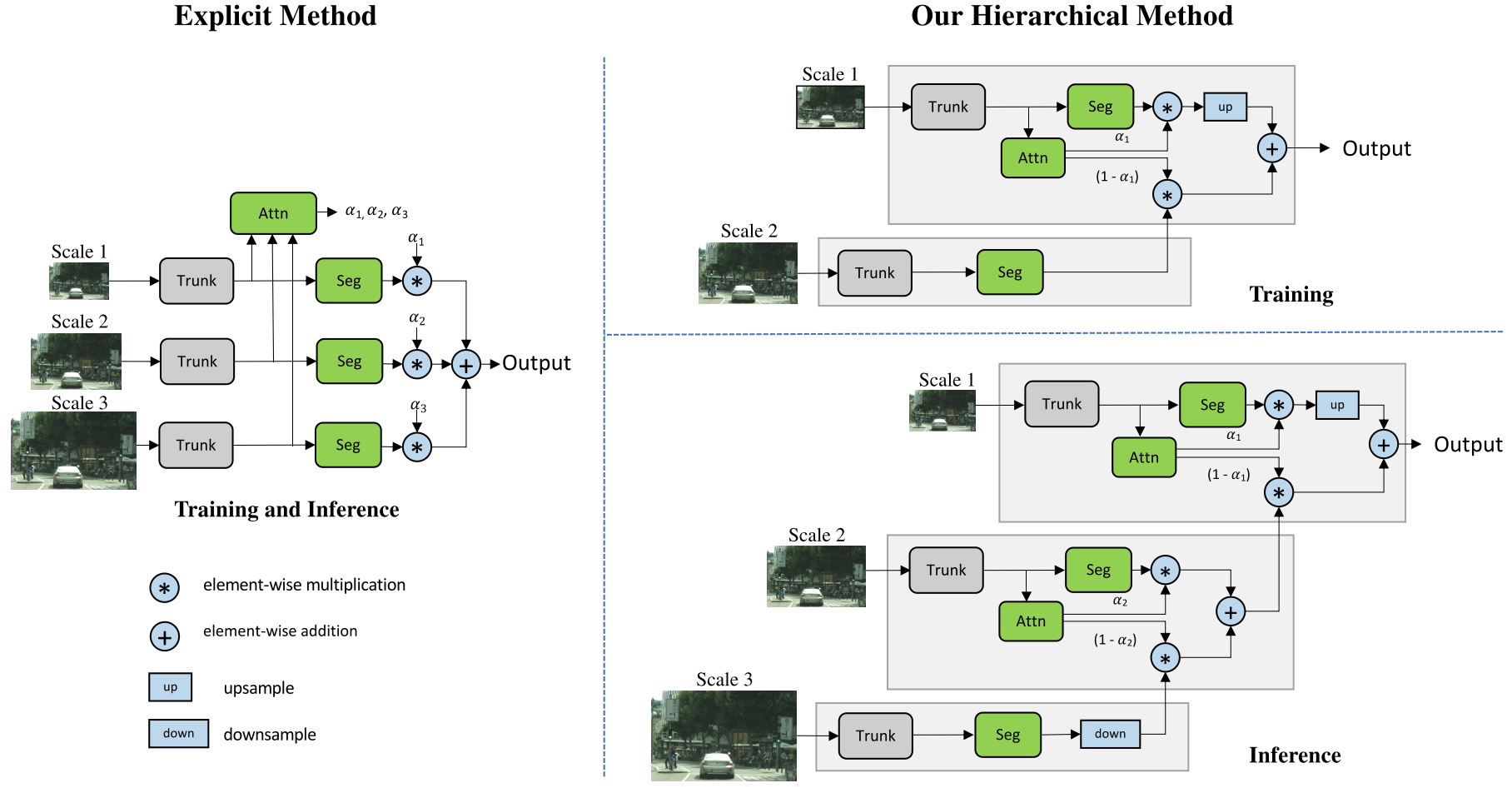}
    \centering 
    \caption{HMA's Architecture visualization from \cite{tao2020hierarchical}. Left panel shows explicit method from \cite{chen2016attention}. Right panel present the HMA's pipeline from \cite{tao2020hierarchical}}
    \label{fig:hma2}
\end{figure}

HMA combines the segmentation result from different-resolution images by a pixel-wise attention mechanism that learns a dense mask for each scale. Using these masks, a weighted aggregation is performed to combine these multi-scale predictions. In stead of learning the masks for each of the fixed scale as in explicit method  \cite{chen2016attention}, HMA uses the hierarchical method which only learns attention map for one adjacent scale and applies it repeatedly to get the mask for other scale. For example, by learning the attention mask for the 0.5x and 1.0x scales, we can generate the attention mask for new scales such as 0.25x and 2.0x. With this hierarchical method, the training efficiency is improved significantly. If we train the model with three scales 0.5, 1.0, 2.0, the training cost is $0.5^{2} + 1.0^2 + 2.0^2 = 5.25$. However, the training cost for two scales 0.5, 1.0 is only $0.5^{2} + 1.0^2 = 1.25$. Hence, the training efficiency is boosted more than 4 times. The HMA architecture and the difference between explicit and hierarchical methods is visualized in figure \ref{fig:hma2}.

\subsection{Domain-Based Batch Normalization}
\label{sec:dbn}

Batch Normalization \cite{ioffe2015batch} has become a standard in computer vision deep neuron network due to its excellent performance in diverse architectures. It is constructed to diminish the internal covariate shifting by standardizing the data with the batch statistics. However, it is assumed that all images belong to the same data distribution. This criterion cannot be guaranteed with the scenario of SMC Challenge 3. To tackle this problem, motivated by the work of \cite{zhuang2020rethinking}, we develop Domain-Based Batch Normalization (DBN) to standardize the data from different distributions and handle the distribution gap between them. For each feature $x_{m}$, the estimated feature $\widehat{x}_{m}$ is calculated as in equation \ref{equ:dbn}

\begin{equation}
\label{equ:dbn}
    \widehat{x}_{m} = \gamma \frac{x_{m} - \mu_{(d)}}{\sqrt{\sigma^2_{(d)} + \epsilon}} + \beta,
\end{equation}
, where $\mu_{(d)}$ and $\sigma^2_{(d)}$ are mean and variance of the domain $d$; $\gamma$ and $\beta$ are the learned parameters from the training process. During training, we feed each mini-batch with the data from one domain and calculated the mean and variance for this domain in current mini-batch. Then, the estimated mean and variance are used to standardize all features in this mini-batch. During testing, a similar process is also conducted to get the standardized features.

%% file: 05-Experience.tex
\section{Experiment Setup and Result }
\label{sec:experience}

\textbf{Dataset Analysis}: The provided dataset consists of 700 real-world images from Cityscapes \cite{cordts2016cityscapes} and 4900 synthetic images from CARLA \cite{dosovitskiy2017carla}, divided into 7 weather conditions (700 images per weather condition): default, clear noon, cloudy noon, cloudy sunset, hard rain noon, mid rain noon, soft rain noon. To get the balance training data, the training set is established by taking the first 600 images of Cityscapes and the first 100 from each weather conditions (exclude soft rain noon) of CARLA. Hence, the training has total of 1200 images including 600 real-world images and 600 synthetic images. The validation set includes the 100 images left of Cityscape and first 100 images of soft rain noon domain from Carla.

\textbf{Data Pre-processing}: We use random crop and random horizontally flip technique to pre-process the data in training stage.

\textbf{Training Detail}: Our proposed models are trained on a single GPU Tesla P100. For the optimizer, we use Stochastic Gradient Descent (SGD) with the momentum 0.9. In addition, the polynomial learning rate policy from \cite{liu2015parsenet} is applied. The batch size is set to be 2.The RMI \cite{zhao2019region} and cross entropy is used as the loss functions. The learning rate is initiated to 0.01 and we used poly exponent of 2.0. The training stage is conducted for 100 epochs.

\begin{figure}[h]
    \centering
    \includegraphics[scale=0.17]{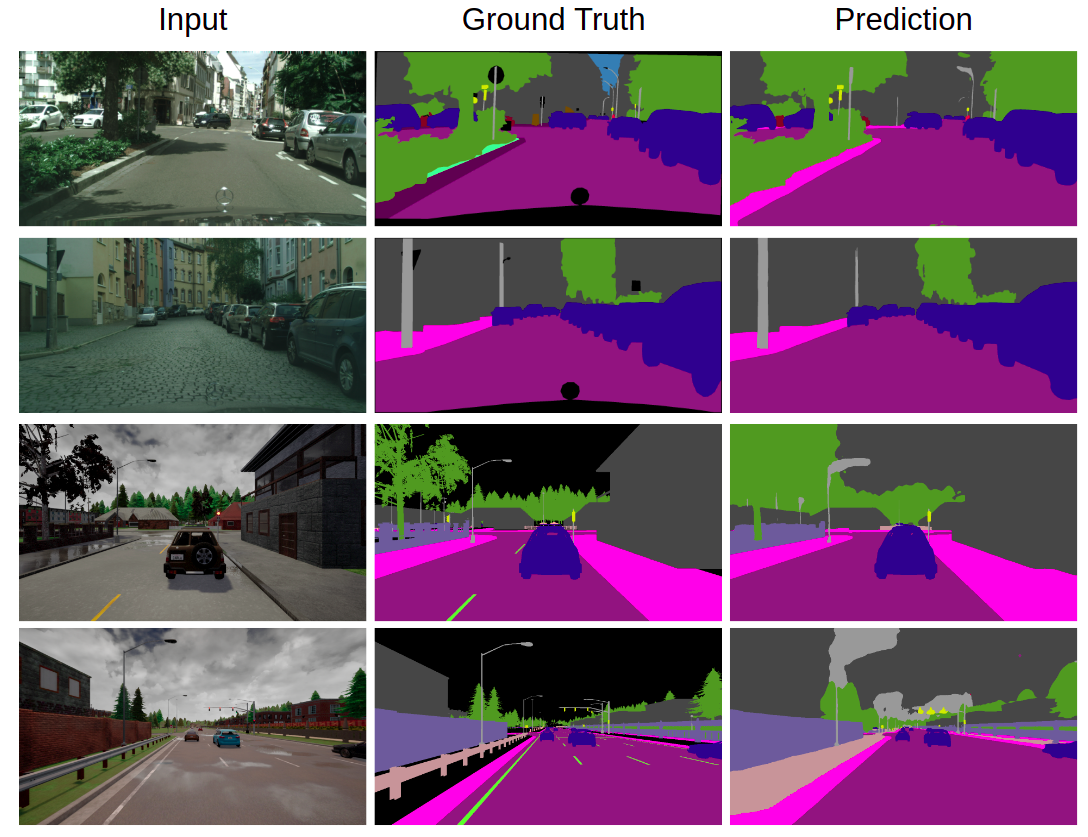}
    \centering 
    \caption{Qualitative Result. First two rows are validation images from Cityscapes and The last 2 rows are validation images from SoftRainNoon of CARLA}
    \label{fig:result}
\end{figure}

\textbf{Result}: At epoch $33^{th}$, we get the result of 81.259 mIoU for the defined validation set. A qualitative result is shown in figure \ref{fig:result}

%% file: 06-Conclusion.tex
\section{Conclusion}
\label{sec:Conclusion}
In this work, first we  applied HRNet \cite{sun2019high} to the coarse pixel-level feature. Then, we enhanced this coarse feature by using OCR \cite{yuan2020object}. This feature was boosted one more time by HMA \cite{tao2020hierarchical}. In addition, we implemented Domain-Based Batch Normalization as domain adaptation technique to reduce the distribution shift from diverse domains. Through our extensive experiments, our approach yields a good mIoU of 81.259. In the future, we plan to apply SegFix\cite{yuan2020segfix} to enhance our model and improve the result even further.